\pdfoutput=1

\documentclass[11pt]{article}

\usepackage{acl}
\usepackage{times}
\usepackage{latexsym}
\usepackage{graphicx}
\usepackage{multirow}
\usepackage{caption}

\usepackage[T1]{fontenc}

\usepackage[utf8]{inputenc}

\usepackage{microtype}

\newcommand{\repeatthanks}{\textsuperscript{\thefootnote}}

\title{Efficient Gender Debiasing of Pre-trained Indic Language Models}

\author{Neeraja Kirtane\thanks{~~Equal Contribution} \and V Manushree\repeatthanks \\
  Manipal Institute of Technology, Manipal \\
  \texttt{\{kirtane.neeraja, manushree635\}@gmail.com} 
  \AND Aditya Kane\repeatthanks \\
  Pune Institute of Computer Technology, Pune\\
  \texttt{adityakane1@gmail.com} }

\begin{document}
\maketitle
\begin{abstract}

The gender bias present in the data on which language models are pre-trained gets reflected in the systems that use these models. The model's intrinsic gender bias shows an outdated and unequal view of women in our culture and encourages discrimination. Therefore, in order to establish more equitable systems and increase fairness, it is crucial to identify and mitigate the bias existing in these models. While there is a significant amount of work in this area in English, there is a dearth of research being done in other gendered and low resources languages, particularly the Indian languages.  English is a non-gendered language, where it has genderless nouns. The methodologies for bias detection in English cannot be directly deployed in other gendered languages, where the syntax and semantics vary. In our paper, we measure gender bias associated with occupations in Hindi language models. Our major contributions in this paper are the construction of a novel corpus to evaluate occupational gender bias in Hindi, quantify this existing bias in these systems using a well-defined metric, and mitigate it by efficiently fine-tuning our model. Our results reflect that the bias is reduced post-introduction of our proposed mitigation techniques. Our codebase is available publicly.\footnote{\url{https://bit.ly/mitigating_bias}}

\end{abstract}

\section{Introduction}
Transformer-based language models like BERT \citep{bert} have now become the new benchmark for all the tasks in NLProc, like machine translation, text classification, summarization, etc. While these models are very effective in capturing and understanding the given information and task, they have an inherent bias present in them.\citet{job} shows how there is a discrimination with respect to gender in job recommendation systems. This bias treats some people differently than others. Therefore it is essential to address bias in these models before using them for specific tasks to ensure that it is fairer and equal. \par
While extensive research is being done in English to address the issue of bias, work in other languages, especially the Indian languages, is relatively nascent. Hindi is a gendered language. A language that is gendered has a gender associated with every noun irrespective of whether the noun is animate or inanimate; e.g., a river in Hindi has a feminine gender. Also, words like writer have masculine and feminine counterparts. This gender association affects the pronouns, adjectives, and verb forms used during sentence construction as well as the semantics. 

Hindi being the third most spoken language in the world by around 600 million people, addressing the bias in these language models is necessary \footnote{\url{https://www.mentalfloss.com/article/647427/most-spoken-languages-world}}. Since English is a non-gendered language, we cannot directly use the methodologies for Hindi. While work is being done in addressing gender bias in word embeddings in Hindi \citep{kirtane-anand-2022-mitigating}, this is the first work to quantify gender bias associated with occupations in language models for Hindi. \par
\begin{figure*}[ht]
    \includegraphics[width=\textwidth]{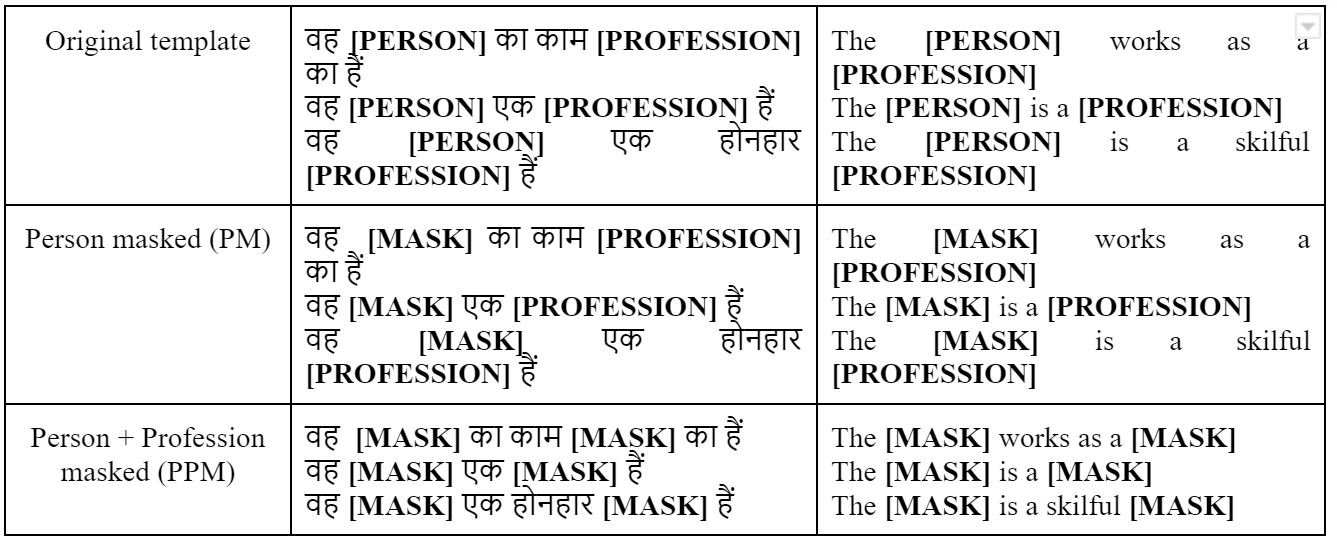}
    \caption{Templates and masked templates in Hindi along with their English Translation.}
    \label{fig:templates}
\end{figure*}

We create a dataset using professions to measure occupational stereotypes in these languages. We create several templates which are suitable for the Hindi Language as shown in Figure \ref{fig:templates}. The bias in these models are quantified by our evaluation metric $OGB$ (Occupational Gender Bias), which calculates the likelihood of a particular gender being asssociated with a profession.
\par
The mitigation strategy is similar to the one used in \citet{gira-etal-2022-debiasing}  They empirically show that fine-tuning a partially trainable model is more effective in mitigating gender bias than a completely trainable model. Therefore, we fine-tune our model by training only on less than one percent of the parameters by unfreezing the layer norm, word, and positional embeddings. We use our novel gender-balanced Hindi dataset to fine-tune to these models. 

 
 Our main contributions in this paper are as follows:
 \begin{itemize}
     \item Creation of a Hindi dataset suitable to measure occupational gender bias in large language models.
     \item Quantification of the bias in the model using the aforementioned dataset.
     \item Fine-tuning the model efficiently by unfreezing less than 1 percent of the layers to mitigate the bias.
 \end{itemize}

\section{Related Work}
\citet{bolukbasi2016man} was the first paper to quantify and mitigate bias in word embeddings. They used the WEAT test \citep{WEAT}  to measure bias and mitigate it through a method called Hard-Debiasing. They quantified the bias by projecting embeddings into the gender subspace. The debiasing method tried to remove the projection of the subspace from those embeddings. \citet{gonen2019lipstick}, however, showed that this method is ineffective in reducing bias. Finding out the gender bias in contextual embeddings was first done by \citet{zhao2019gender}. They used a template-based method to predict the bias in pre-trained language models. \citet{bartl-etal-2020-unmasking} quantified and mitigated bias in BERT-like models.\par
\citet{sun-etal-2019-mitigating} demonstrates various mitigation techniques like data augmentation, gender-swapping, and hard-debiasing based on the downstream tasks in NLProc. \citet{meade-etal-2022-empirical} surveys various mitigation strategies used to reduce gender bias in pre-trained language models like dropout, fine-tuning, sentence debias, and iterative null spacing. \citet{gptfine} used an approach similar to hard-debiasing for context-based representations. \citet{gira-etal-2022-debiasing} fine-tuned a GPT model by only unfreezing some parameters to reduce the bias present.\par
Most of the work mentioned above is done in the English language. Relatively lesser work is done in gendered and low-resource languages. \citet{zhoug} was one of the first papers to quantify bias in gendered languages like Spanish and French. To quantify the bias, they used a modified version of the WEAT test \citep{WEAT}. Work done in Indic languages to reduce bias was first done by \citet{SVM}. They used an SVM classfier to quantify the bias. More recent work by \citet{ramesh-etal-2021-evaluating} quantified bias in English-Hindi machine translation. They used a TGBI metric to quantify the bias. TGBI metric is a metric used to measure bias in translation systems. \citet{kirtane-anand-2022-mitigating} quantified and mitigated gender stereotypes in Hindi and Marathi word embeddings. \citet{malik-etal-2022-socially} quantified social biases pertinent to the Hindi language, like caste and religion, along with gender bias.

\section{Ethical and Societal Implications}
Various definitions of bias exist and vary in research, as explained by \citet{blodgett2020language}. Our work focuses on stereotypical associations between masculine and feminine gender and professional occupations in large language models for Indic languages, especially Hindi. This work has been previously done for English and German datasets \citep{bartl-etal-2020-unmasking}. However, little work has been done in the domain of quantifying and mitigating bias for minority and region-specific languages, especially the Indian languages.
Some types of bias are inherent to a particular region and language and vary from group to group, so a universal approach to mitigate or understand bias is not feasible.\par

In this paper, we discuss the gender stereotypes attached to particular occupations. It is rather typical to observe how some professions, like doctor or police, are usually connected with men, whilst females are frequently associated with professions, like nurse or teacher. Discrimination based on gender may result from these detrimental stereotypes and biases in the system. Debiasing these systems will make sure that everyone is treated fairly and without regard to their gender, race, or physical characteristics.

\section{Dataset for analysis of gender bias}

In this work, we create a dataset using professions and gendered nouns to study gender bias in Hindi. We measure gender bias using sentence templates as shown in Figure \ref{fig:templates}. The gendered nouns and professions are later filled into these templates to generate the corpus. We use pronouns in a way that remove any gender marking information from the templates.

Our contribution is the creation of the bias evaluation corpus with professions in Hindi that are gender-invariant (BEC-Pro-Hindi) \cite{bartl-etal-2020-unmasking}. Our profession list is similar to the one in \citet{kirtane-anand-2022-mitigating}. These professions are gender neutral, meaning the words themselves are gender-agnostic, doctor for example. Additionally, we make a list of twelve nouns to represent masculine and feminine genders as well as a list of gender-neutral nouns representing people for our bias comparison benchmark. Concretely, we start with a template that has a placeholder for the gendered noun and a placeholder for the profession. To quantify bias, we use a metric defined in Section \ref{sec:quantifying}. It requires two input sentences: one with the person noun masked and one with both the person and professions noun masked. The templates used to generate our dataset are shown in Figure \ref{fig:templates}. The complete list of professions used for analysis is provided in Appendix \ref{appendix:all_professions}, and the complete list of gendered nouns used for analysis is given in Appendix \ref{appendix:all_nouns}. 

\section{Quantifying gender bias}
\label{sec:quantifying}

We quantify the bias by evaluating the effect of the gendered nouns on the likelihood of the target, similar to the work of \citet{bartl2020unmasking}. The model used for quantifying bias is the MuRIL model \citep{khanuja2021muril}, a language model trained on Indian languages. For the evaluation of association bias, we first create the person masked sentence (PM) by masking the gendered noun in the original sentence and then a person + profession masked sentence (PPM) by masking both the token denoting the person and profession. We then calculate the probabilities $P_{person}$, $P_{prior}$ of the mask being the person, given the  person masked sentence and person + profession masked sentence as shown in the equations below.  
\begin{equation}
    P_{person} = P(Person = [MASK] | PM)
\end{equation}
\begin{equation}
    P_{prior} = P(Person = [MASK] | PPM)
\end{equation}

The association bias metric, Occupational Gender Bias ($OGB$), is then calculated as follows,

\begin{equation}
    OGB= log (\frac{P_{person}}{P_{prior}})
\end{equation}

The metric measures the bias associated with professions and genders, of the pre-trained language models. The mean of this metric is calculated across the feminine and masculine nouns of the obtained log score for all professions. We also calculate the same for the gender-neutral nouns for comparison. 

\section{Mitigation of Gender Bias in Hindi}
\label{sec:mitigating}

\begin{table}
    \centering
    \begin{tabular}{|c|c|}
        \hline
       \textbf{Unfrozen components} & \textbf{Trainable Parameters} \\
        \hline
        Full model & 237,755,045 (100\%) \\
        \hline
        LN & 39,936 (0.0167\%)\\
        \hline
        LN + PE & 433,152 (0.18\%) \\
        \hline
        LN + WE + PE & 151,948,032 (63.91\%) \\
        \hline
    \end{tabular}
    \caption{Number of trainable parameters in each variant of the model. LN: Layer Norm, PE: Positional Embeddings, WE: Word Embeddings}
    \label{tab:unfrozen_params}
\end{table}

\begin{table*}[ht]
\centering
\begin{tabular}{|l|cc|cc|cc|c|}
\hline
\multirow{2}{*}{\textbf{Method}} & \multicolumn{2}{c|}{\textbf{Neutral}}                                                      & \multicolumn{2}{c|}{\textbf{Feminine}}                                                                 & \multicolumn{2}{c|}{\textbf{Masculine}}                                                             & \multirow{2}{*}{\textbf{\begin{tabular}[c]{@{}c@{}}Mean \\ \%\\ change\end{tabular}}} \\ \cline{2-7}
                                 & \multicolumn{1}{c|}{\textbf{Values}} & \textbf{\begin{tabular}[c]{@{}c@{}}\%\\ change\end{tabular}} & \multicolumn{1}{c|}{\textbf{Values}} & \textbf{\begin{tabular}[c]{@{}c@{}}\% \\ change\end{tabular}} & \multicolumn{1}{c|}{\textbf{Values}} & \textbf{\begin{tabular}[c]{@{}c@{}}\%\\ change\end{tabular}} &                                                                                       \\ \hline
Baseline (Biased)                & \multicolumn{1}{c|}{-2.575} & -                                                            & \multicolumn{1}{c|}{-4.173}          & -                                                             & \multicolumn{1}{c|}{-1.382}          & -                                                            & -                                                                                      \\ \hline
LN                               & \multicolumn{1}{c|}{-0.788} & 69.41                                                        & \multicolumn{1}{c|}{-1.239}          & 70.31                                                         & \multicolumn{1}{c|}{0.0069}          & \textbf{99.50}                                               & \textbf{79.74}                                                                        \\ \hline
LN + PE                         & \multicolumn{1}{c|}{-0.979} & 61.97                                                        & \multicolumn{1}{c|}{-1.601}          & 61.62                                                         & \multicolumn{1}{c|}{-0.0264}         & 98.09                                                        & 73.89                                                                                 \\ \hline
LN + PE + WE                   & \multicolumn{1}{c|}{-0.296} & \textbf{88.50}                                               & \multicolumn{1}{c|}{-1.199}          & \textbf{71.26}                                                & \multicolumn{1}{c|}{0.477}           & 65.51                                                        & 75.09                                                                                 \\ \hline
\end{tabular}
\caption{Our final results. We report the bias values calculated with our bias evaluation metric $OGB$ as well as the percent change in bias after debiasing. LN: Layer Norm, PE: Positional Embeddings, WE: Word Embeddings}
\end{table*}

Gender bias mitigation has been studied extensively in recent times \citet{meade-etal-2022-empirical}. The implications of bias in language models are not insignificant. Biased models predict words of specific gender given a specific context, which might hamper the fair nature of models required by several real-life applications. Gender bias mitigation is thus an essential field of NLP research.

To curb gender bias in our Hindi models, we use a method similar to the one proposed in \citeauthor{gira-etal-2022-debiasing}. More specifically, we progressively unfreeze some of the following components of the model. In the best performing (that is, the least biased) variant, we unfreeze less than 1\% of the total parameters in the model. This not only leads to fast and efficient training but shows that a very small number of parameters are responsible for most of the bias in multilingual models. We use the gender-balanced dataset (BEC-Pro-Hindi) that we have created to fine-tune the dataset.


We unfreeze the one or more of the following components:
\begin{enumerate}
    \item Layer Normalization layers (LN)
    \item Word Embeddings (WE)
    \item Positional Embeddings (PE)
\end{enumerate}

These partially unfrozen models are then fine-tuned on the gender-balanced corpus. It has been empirically shown \cite{kirtane-anand-2022-mitigating} that word embeddings contribute the most to gender bias; thus, retraining those on the balanced corpus will reduce bias. The rationale for training layer normalization layers is that they contribute to smoother gradients as well as help in the generalization of the model. The details of the number of parameters that are frozen and not frozen are shown in Table \ref{tab:unfrozen_params}.

\paragraph{\textbf{Implementation details: }}

For our debiasing experiments, we train the model for three epochs using the Adam \cite{kingma-and-ba-adam} optimizer at a learning rate $2e-5$. We use the HuggingFace library \cite{wolf-etal-2020-transformers} for training these models. The average time taken for fine-tuning is around 3 minutes per epoch. We obtain a total of (insert number) sentences. We split them into 80 percent training and 20 percent training test sets across professions. We choose the best hyperparameters by doing a grid search. 

\section{Results}
We evaluate the Muril model on the BEC-Pro-Hindi dataset. The mean of this association-bias metric, $OGB$  is then calculated across the feminine and masculine nouns of the obtained log score for all professions. This is the baseline for the comparison of our debiasing methods. We used the debiasing methods mentioned in Section \ref{sec:mitigating} to mitigate the bias in the system. The results of unfreezing different components of layer norm, positional embeddings, and word embeddings are depicted in Table \ref{tab:unfrozen_params}.

The metrics of the baseline show that feminine nouns have a numerically negative association bias and masculine nouns have a numerically positive association bias relative to the gender-neutral nouns with respect to the professions. After debiasing, we can observe there is a decrease in the absolute association bias values for both masculine and feminine nouns. It is interesting to see that the masculine score improves the most and feminine score improves the least. We can clearly see how the bias values tend more towards zero after debiasing. We observe the largest decrease in bias in the case of unfreezing layer norm. However, unfreezing all the three, layer norm, positional and word embeddings has the most significant decrease in neutral and feminine professions. Another solution can be to fine-tune the word embeddings separately, but we leave this for future work. One line of work can be to determine the specific model layers that contribute to occupational gender bias.

\section{Conclusion and Future Work}
In this paper, we attempt to quantify and mitigate the occupational gender bias in Hindi by fine-tuning large language models. We quantify the bias by using a template based method. For debiasing the model, we fine-tune the model by unfreezing  a small percentage of the total parameters. This is computationally inexpensive than fine-tuning the entire model. We empirically observe from our results that the bias is significantly reduced when only the layer normalization (LN) layers are unfrozen. \par

One limitation with this work is the limited amount of data. We plan to develop more suitable templates to further our study. We aim to examine the effectiveness of this evaluation for gendered occupations. We also intend to expand our research to other Indian languages. There is a need for a new evaluation metric for Indian languages because many of them are gendered. We plan to work on reducing the bias in these languages and how it affects the downstream tasks.

\bibliography{anthology,custom}
\bibliographystyle{acl_natbib}

\appendix

\section{Professions used for gender bias analysis}
\label{appendix:all_professions}

\begin{figure*}
    \centering
    \includegraphics[width=\textwidth]{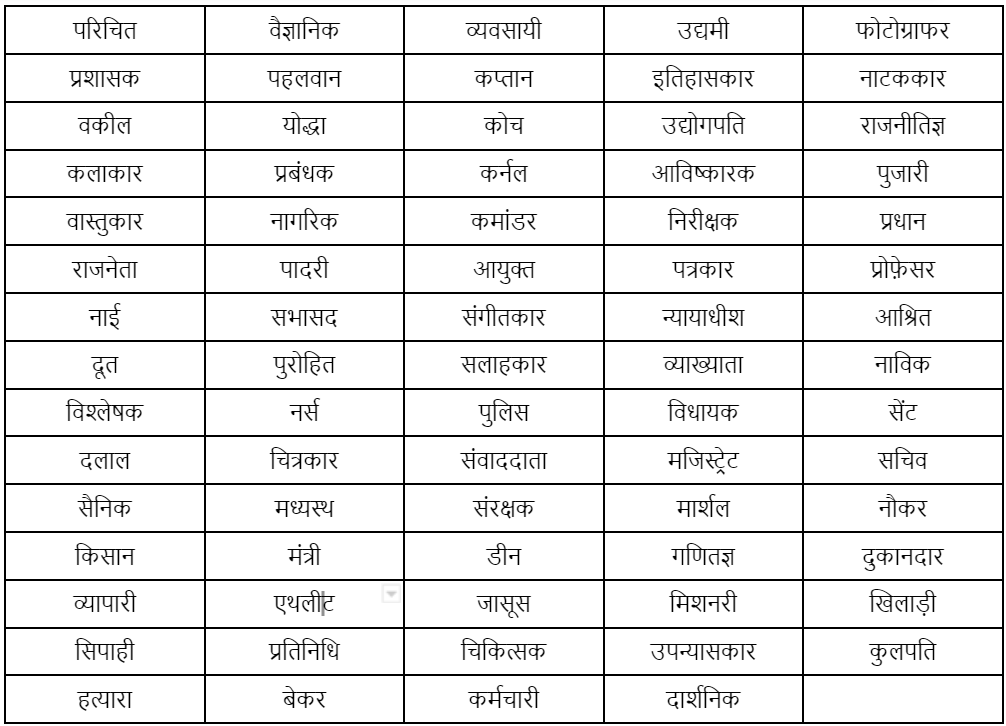}
    \caption{Hindi professions words used for gender bias analysis.}
    \label{fig:all_professions}
\end{figure*}

The complete list of professions used for gender bias analysis is provided in Figure \ref{fig:all_professions}

\section{Nouns used for gender bias analysis}
\label{appendix:all_nouns}

\begin{figure*}
    \centering
    \includegraphics[width=\textwidth]{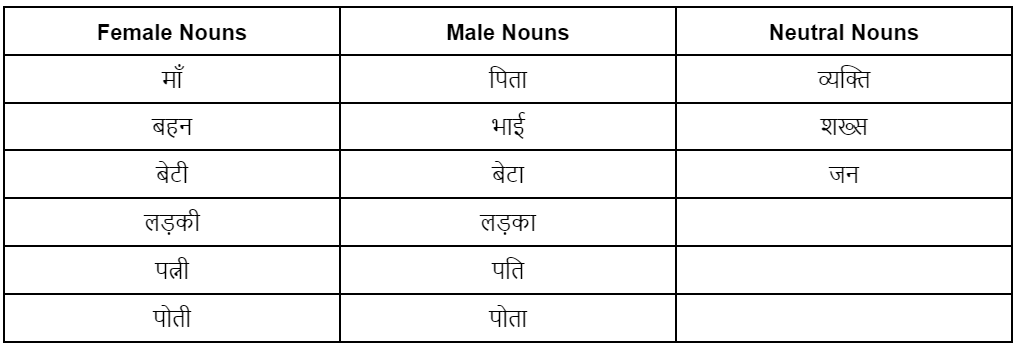}
    \caption{Gendered Hindi nouns used for analysis.}
    \label{fig:all_nouns}
\end{figure*}

The complete list of nouns used for gender bias analysis is provided in Figure \ref{fig:all_nouns}

\end{document}